\documentclass{article}
\usepackage[utf8]{inputenc}

\title{SenseNet: 3D Objects Database and Tactile Simulator\\ http://sensenet.ai}
\author{Jason Toy}
\date{December 2017}

\usepackage{graphicx}
\graphicspath{ {images/} }

\begin{document}

\maketitle
\section{Abstract}

 The majority of artificial intelligence research, as it relates from which to biological senses has been focused on vision. The recent explosion of machine learning and in particular, deep learning, can be partially attributed to the release of high quality data sets for algorithms from which to model the world on. Thus, most of these datasets are comprised of images. We believe that focusing on sensorimotor systems and tactile feedback will create algorithms that better mimic human intelligence. Here we present SenseNet: a collection of tactile simulators and a large scale dataset of 3D objects for manipulation. SenseNet was created for the purpose of researching and training Artificial Intelligences (AIs) to interact with the environment via sensorimotor neural systems and tactile feedback. We aim to accelerate that same explosion in image processing, but for the domain of tactile feedback and sensorimotor research. We hope that SenseNet can offer researchers in both the machine learning and computational neuroscience communities brand new opportunities and avenues to explore.

\section{Introduction}

The majority of machine learning research has been focused on vision and image processing. Visual processing is critical to the survival of most species, especially humans. Humans use vision to do tasks such as driving, watching movies, reading, learning, playing sports, and more. Without vision we would not be able to complete most of these tasks. Furthermore, digital cameras have allowed a way for computers to capture and process visual data in a comparable way to humans and so it makes sense that the majority of machine learning is focused on vision processing. That given, we believe that while this focus on vision has brought about many advances in machine learning and computer vision algorithms, it is fundamentally the wrong direction towards general  artificial intelligence. While not every living animal has vision, every single living animal has some sense of touch. Touch is not confined to the fingers, but all over the skin.
Almost all the recent breakthroughs in computer vision models has come from deep neural network architectures. These networks use base units such as convolutional layers, max pooling, and transpose layers to transform the inputs in a loose way that mimics retina neurons. These architectures are mostly static, with the image inputed into the model, the model then converts the image into neural network inputs that get transformed and modified through the layers until the final layer  which typically outputs a probability distribution. These models do not really take into account time and motion, yet our senses are always occurring over time and with motion. For example, in human eyes motion is critically involved. The eye is constantly moving multiple times a second in saccades, scanning what appears on the retina to reconstruct the image into a new image that appears as a stable image in our minds. This motion process happens over time in both a conscious and subconscious way as our eyes gaze over a scene.  This same movement phenomena is fundamental to tactile recognition. Image your eyes closed and placing your index finger on a coffee cup. Without any movement, you can't tell that it is a coffee cup, but if you move your finger over the cup, your brain builds an internal model of the object and you recognize the features such as the curve of the cup and the mouth of the lid. After this movement, you are able to recognize that the object is a coffee cup.

Demis Hassabis recently stated that to build better AIs, we must turn to neuroscience\cite{demis}.
If there is a single core learning algorithm that can learn over the different sensory inputs, then time and sensorimotor coordination is a fundamental aspect of that algorithm.
SenseNet is meant to be used as a research framework for sensorimotor systems and tactile feedback for computational neuroscience and machine learning researchers.
We hope with the release of SenseNet as an open source project to the world, more robust and sophisticated models can be explored and built, resulting in better end user applications and understandings of intelligence.
Our overarching goal in releasing SenseNet is to create that initial spark of research and exploration in sensorimotor systems and tactile feedback that ImageNet did for the computer vision field.

\section{SenseNet and related Datasets}

Over the past decade there has been several large datasets released to aid in the research and exploration of computer vision algorithms.  ImageNet, a database of over 1000 classes totaling over 15 million images is notable for accelerating the progress of computer vision models\cite{imagenet}. Classification error rates of images from belonging to the ImageNet dataset started out at 71.8\% and have recently obtained an accuracy rate of 97.3\%. Many new types of deep neural network architectures were released such as AlexNet\cite{alexnet} while managing to achieve a top 5 test error rate of 15.4 percent. As of now, ImageNet has over 15 million images ranging from common objects to animals, to rooms and vehicles.  Over 70% of ImageNet's classes are animal species and so many of their classes are not modeled in SenseNet. ShapeNet is dataset of over 3 million 3D CAD models of objects. ShapeNet's stated purpose is to aid in scene understanding of 2D images with the assistance of their 3D models\cite{shapenet}. Their dataset has a wide variety of objects including rope, fungus, grass, and cheese. SenseNet currently only focuses on hard body objects that can be modeled accurately in a computer simulation. Objects such as fungus have soft bodies and have too many properties that cannot currently be modeled in a general way in current simulations. The Yale-CMU-Berkeley (YCB) Object and Model set is small dataset of 77 objects categorized into 5 classes meant for robotic grasping and manipulation research \cite{ycb}. The dataset has both 3D models and a set of physical objects that they will mail to you. YCB's 3D models are typically used with Gazebo, a robotics simulator. Most of these datasets are focused primarily on image applications where the modality that the machine learning models experience the dataset is through computer vision. SenseNet, on the other hand was created explicitly for researching and experimenting with models that deal extensively with tactile feedback.

\section{Properties of SenseNet Dataset}

Most of the other datasets are either very small or are large and link to the original files somewhere else on the internet. Typically this is due to copyright reasons and datasets being too large to host. Originally we investigated the possibility of indexing all the 3D content available online, but we found the data to be too messy to manually clean and organize by hand. Instead we opted to start off with parametric design. Almost all of our objects are designed and generated by code with multiple parameters to control width, height, protrusions, radius, and more. In this way we are able to programmatically control the overall look and shapes of the data while creating variety in the dataset.

\includegraphics[scale=0.3]{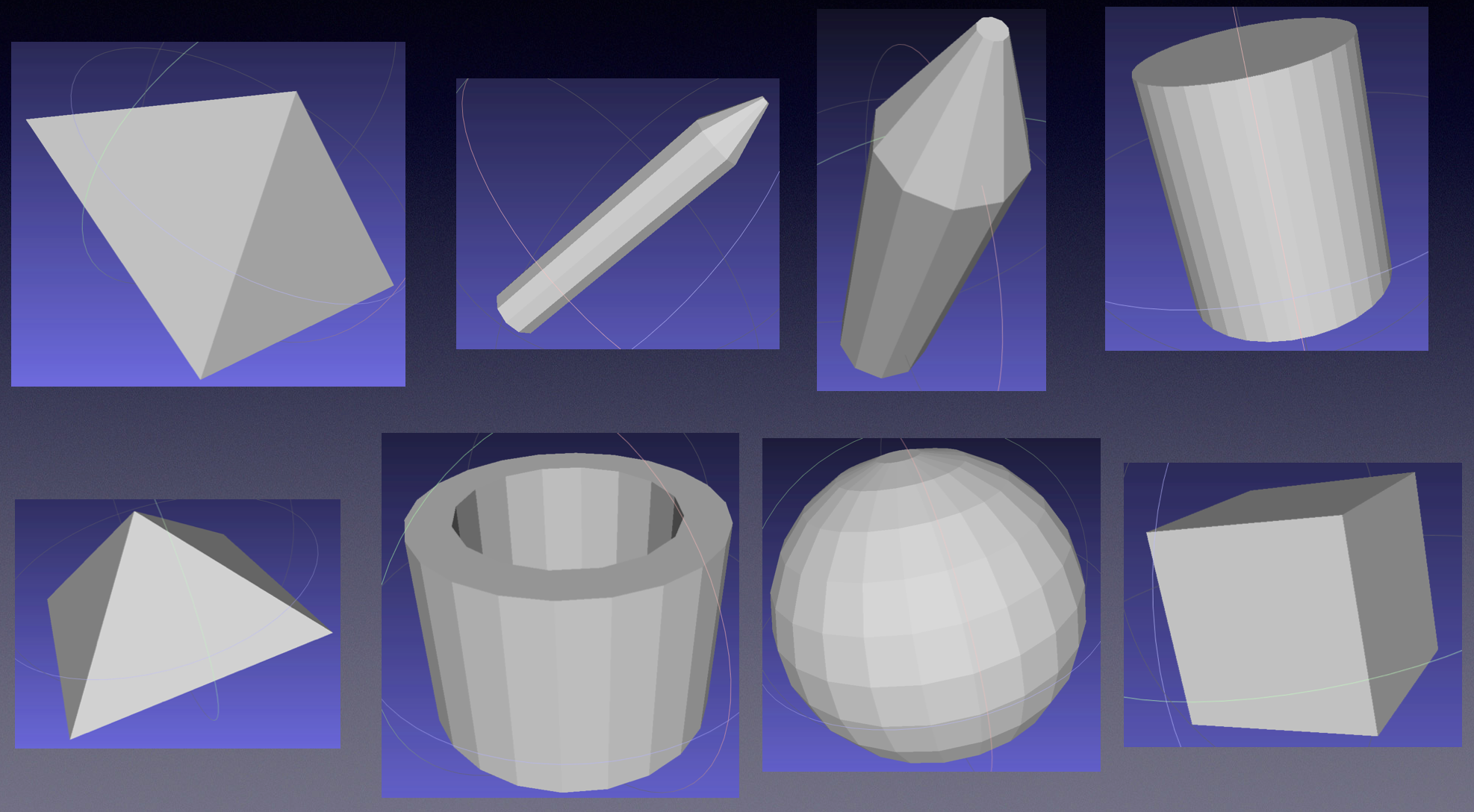}

SenseNet aims to be the most comprehensive and largest dataset of 3D touchable objects. As the dataset grows over time, we will introduce more manually modeled objects to increase variety. The initial dataset consists of 20 tools that are typically manipulated with hands. Similar to ImageNet, we categorize our classes into a hierarchy based off of WordNet\cite{wordnet}. We aim for each class to have 500 to 1,000 objects. We do not need as many examples as vision datasets because we can view our objects from multiple angles and orientations creating copious variations of our data. Some of the current object classes are cups, cubes, missiles, vases, pyramids, cylinders, plates, bowls, pencils, bottles, spheres,spoons, and hammers. The dataset is currently focused on solid objects that can have an accurate single format that can be presented in both the digital world and real physical world. All files in the dataset are represented as both OBJ files and Standard Tessellation Language (STL) files. STLs and OBJs files are native file formats for CAD systems and thus are a industry wide standard, free, and open. STL files and OBJ are both used as standard file formats for 3D printing. Since we use the STL and OBJ file formats, all of our objects can be 3D printed without any modification, thus creating a canonical object definition in both the real and digital world. We include handheld size models, miniature models, and large scale models.
The initial dataset is focused on solid hard bodied objects with rigid edges and defined boundaries. Datasets like ShapeNet include a wide variety of 3D models like fungus,pizza, and roses. While we could have implemented those objects, their physical properties can not be modeled with enough precision in current simulations.  Pizza, while a solid object, is mushy,spongy, and composed of many other objects like tomato sauce,dough,etc. STL and OBJ files describes only the surface geometry of three-dimensional objects, and cannot represent other properties such as mass, weight,texture, viscosity, color, etc.
We focused on handheld tools because by definition, these are items that the human hand is used to manipulating.
We are constantly adding new objects to the dataset and expect the dataset to reach over 1000 classes with over a million objects.
Our intent is to eventually create a database of all common objects.  We suspect that a new format will need to be created to accurately capture all the different properties of the objects we intend to model.

\section{Touch and Sensorimotor Neurons}

Skin acts an interface between the internal and external world of organisms.  The purpose of skin has two purposes, to act as a barrier for the organism to protect from things such disease, chemicals, parasites, and to allow touch information in for the organism to process. Human skin is known to have four types of mechanoreceptors that can measure temperature, stretching of the skin, low and high frequency vibrations, points, and edges.  Each of these receptor types sends information to the spinal chord on dedicated paths, but by the time it reaches our conscious access, it has been processed and blended in our brains. Moreover, touch information is subconsciously combined with other inputs such as proprioception.
The tips of the fingers of the human hand has the highest concentration of merkel cells, approximately 2000 cells per square centimeter.
It is generally accepted that merkel cells exists ubiquitously in vertebras including humans and it is where the SenseNet simulator is currently focused on.
Living organisms have neurons that process sensory data and motor neurons that when fired, produce movement in the organism. Sensorimotor systems refer to this coupling of sensory and motor neurons that work together to produce movement and new sensory input from interacting with the environment.

\includegraphics[scale=0.15]{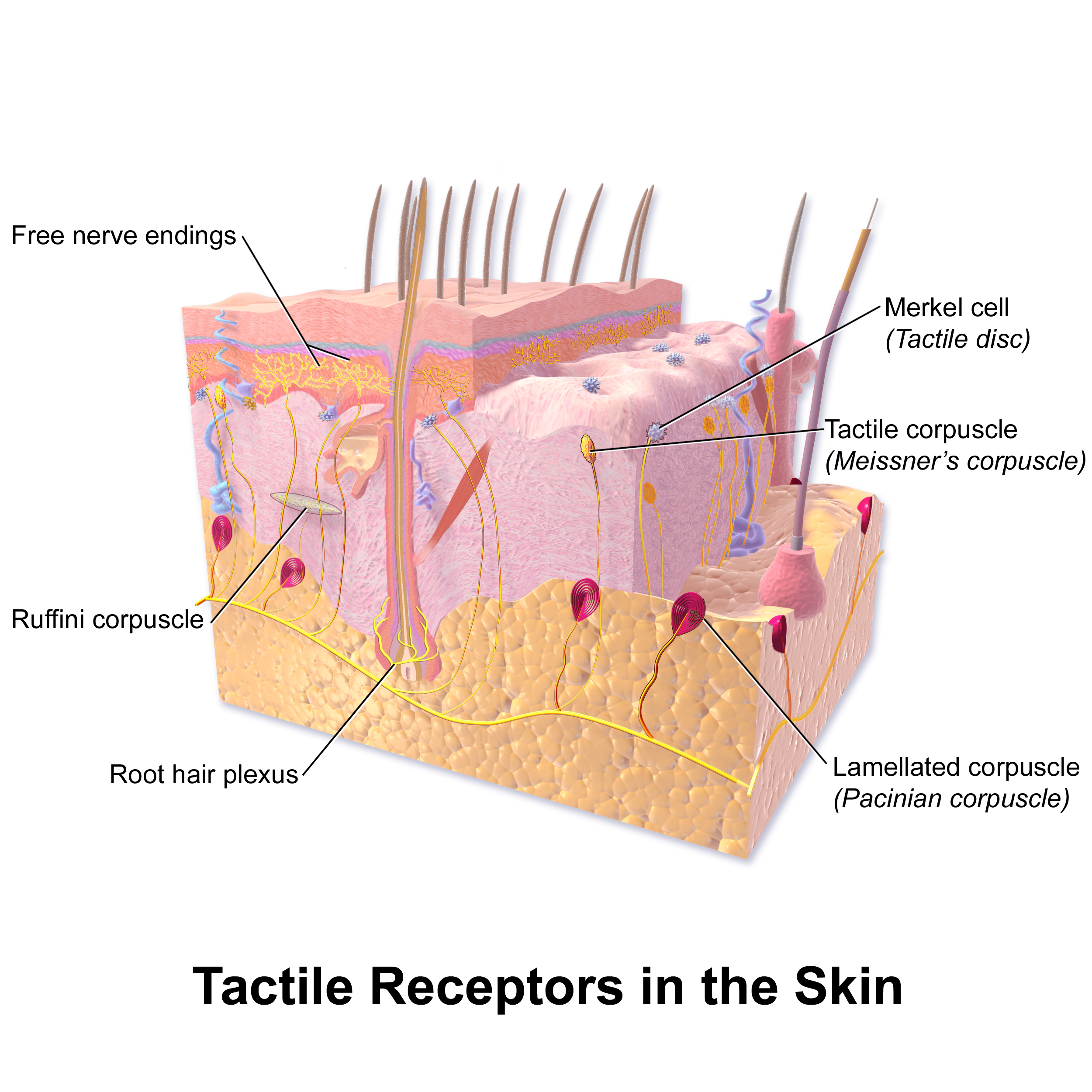}
wikipedia\cite{skinpic}

\section{Robotics and sensorimotor systems}

Robotics is a key area of research in bringing physical mobile computers to the real world.  Robots must learn to navigate the world in the same space that we occupy and are therefore bound to the same laws of physics as us. Robotics is an active area of research where researchers are studying topics such as action planning, locomotion, and kinematics. Developing tactile touch sensors has proven difficult,expensive, and elusive to build onto robots and thus far very few robots in the commercial world have tactile feedback. The vast majority of these robotics systems use cameras as their main sensory input to the real world.
Artificial Intelligence systems have typically been built as abstract computer programs that process sensory data as tacked on modules. All intelligent systems we know of have biological bodies and as more experiments are done, we find that the body and movement of the body influences the mind as well. There has been a growing trend of researchers focusing on routines for interacting with the environment instead of internal representations used for abstract thought and this approach has been collectively named embodied cognition. The embodied cognition hypothesis states that an unknown, but potentially large fraction of animal and human intelligence is a direct consequence of the perceptual and physical richness of our environment, and is unlikely to arise without it. There has been several interesting studies where researchers have tried to model different parts of sensorimotor systems. One group of researchers tried to encode proprioceptive inputs into the brain with a robot simulation and came to the conclusion that what we know about proprioception is contradictory and that we don't know enough about how proprioception works\cite{proprio}. There is recent work by Abduh and Hawkins to understand and model how the brain differentiates external sensory data versus sensory data caused by movement. They hypothesis that a single neural mechanism can learn and recognize both types of sequences\cite{numenta}.

\section{Touch and Motion in machine learning research}

The majority of current research in machine learning has been focused on vision.  It is our hypothesis that there is a core learning algorithm in the human brain that works across all the modalities of sensory input including touch and vision. We believe that there is a lack of machine learning research in tactile feedback for several reasons. Firstly, vision is the easiest sense for us to research and it seems like the most important sensory input. In terms of the amount of information you can process at one moment, vision has wider "bandwidth" and can see far away while touch has smaller bandwidth and can only sense local objects in proximity to the organism.  Motion does not seem to be a requirement for vision, but that is false, we know the eyes move in saccades several times a second reconstructing visual input into a stable image in our minds. Motion is constantly happening at the subconscious level of visual processing, while the algorithms written today do not take into account motion and time at the fundamental level. There are some successful attempts to add motion-like capabilities to computer vision with algorithms called attention mechanisms\cite{attention} that work by focusing on certain regions of an image in higher resolution and adjusting the focal point over time.  While these attention mechanisms have been deployed successfully, they have been designed to work on top of existing convolutional neural network architectures and not built as a new core building block of neural network algorithms.
There are branches of deep learning research that focus on sequential data such as Recurrent Neural Networks (RNNs) and Long Short Term Memory (LSTM) networks and they do have successful applications in domains such as machine translation, speech recognition, and natural language processing. These RNNs are mostly applied to textual data.  There have been attempts at combining RNNs with CNNs in a unified architecture called Convolutional, Long short-term memory, fully connected deep neural networks (CLDNN)\cite{cldnn}, but have thus far not seen much success beyond research applications. Second, we do not have a strong enough unified understanding of how the sensorimotor system works in order to turn it into an algorithm.
We outlined earlier some interesting areas of research in sensorimotor research but no grand theory has appeared yet. The research and scientific process cannot be sped up, but awareness of the problem can increase awareness and interest in the subject matter.
Third, Even if we did have a unified understanding of sensorimotor systems, there is no simple way to test these theories in the context of machine learning research. Building physical robots and the sensors required is expensive and prohibitive. This could be built in software simulations, but to do so would need development of a sophisticated simulator that mimics physics, skin, and touch in a way that is similar to our own world and yet computationally feasible. The cost and burden of developing this is cumbersome and takes away from the core of researching new sensorimotor theories. Ideally there would be a standard simulator to test hypothesis on so as to be able to compare and contrast work.

\section{SenseNet Simulator}

Providing a 3D dataset of objects is not enough if you want to conduct sensorimotor systems and tactile feedback research. A simulator to load and manipulate those objects is needed as well. Thus, as part of the release of this paper and dataset, we have included a python simulator for researchers to run their simulations on.  We have built a layer upon the bullet physics engine\cite{pybullet}. Bullet is a widely used physics engine in games, movies, and most recently robotics and machine learning research.  It is a realtime physics engine that simulates soft and rigid bodies, collision detection, and gravity. We include a robotic hand called the MPL that allows for a full range of motion in the fingers and we have embedded a touch sensor on the tip of the index finger that allows the hand to simulate touch.

\includegraphics[scale=0.35]{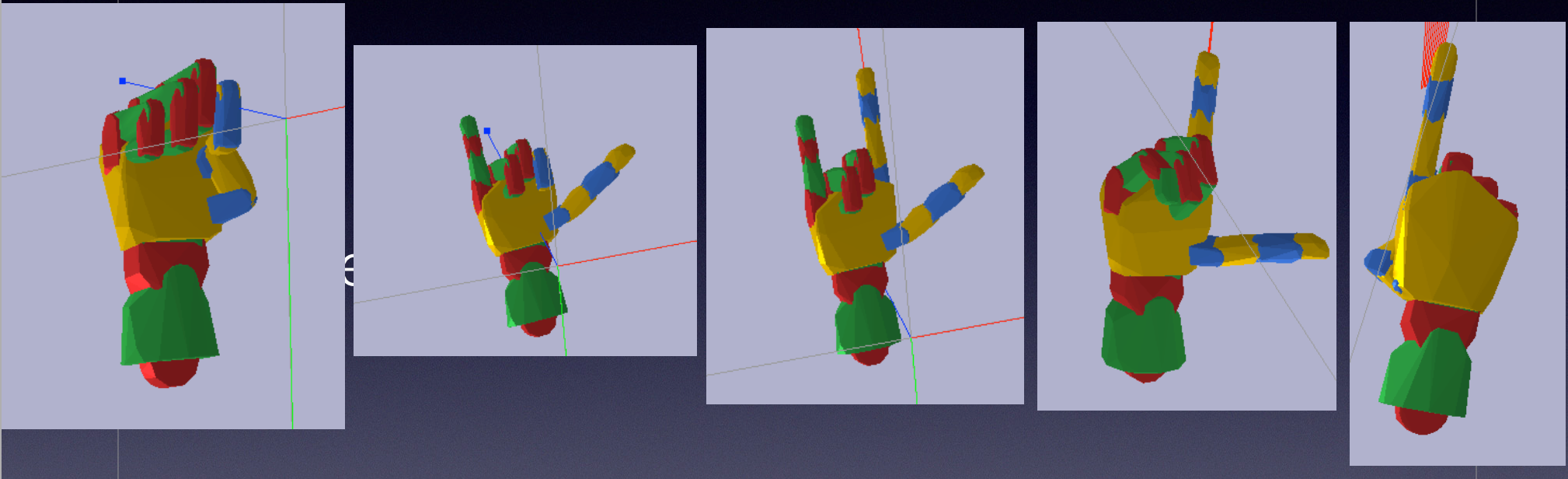}

SenseNet robotic hand gestures

\section{SenseNet API}

Reinforcement Learning (RL) is a branch of machine learning inspired by behaviorist psychology that is concerned with how software agents should take action in an environment to maximize some reward. There have been many recent examples combining deep learning and reinforcement learning to surpass previously thought to be undoable by AI, such as Google's AlphaGO model.  Since reinforcement learning algorithms must deal with an environment, working over time is a hard requirement for these algorithms, and as such, a better fit for sensorimotor neuron simulations compared to other machine learning techniques. We developed an API that follows the recent trend of reinforcement learning algorithms where we provide the simulator as an abstracted environment object in python. In this way, as we add new simulation versions, they can easily be swapped out and the agent's code does not need to modified.
Time is implemented as discrete steps where the programmer can explicitly choose where in their code to move forward in time. The step function is provided as part of the environment object to move forward in 1 discrete step in time. Each step through the environment returns back observations from the sensory inputs which can be used by the machine learning agent to process. The observation data returned in our initial environment is touch data at the tip of the index finger. The data is returned back as a 40x40 array representing width and height of  the touch apparatus. The possible values of the array are discrete values of  0's and 1's.  1 representing a touch and 0 representing no touch. We may implement continuous values for the touch apparatus in future versions. Every API endpoint that provides sensor data is an the form of multi level arrays using numpy\cite{numpy} to remain compatible with the scientific community.
 Our API was designed to be compatible with OpenAI's gym API, a recent collection of various environments to explore different reinforcement learning algorithms\cite{openai}. This means that the majority of software written to work with OpenAI's gym will work on SenseNet.
The SenseNet dataset and simulator is not required to be used in conjunction with reinforcement learning, but it provides a familiar and fast way to test out new hypothesizes. SenseNet's data and simulation could be used in more traditional areas of machine learning such as pure classification and regression.

Our initial environment provides a robotic hand where each finger can be independently controlled with a total of 27 actions that can be taken at a single time step. These actions are for moving each joint on each of the fingers and moving in the x,y,z access for the base of the hand. For each time step through the environment, only 1 action can be taken, currently you cannot combine multiple actions in a single time step.

The software is extendable so that researchers can plug in their own environment or hand model as well. Our current hand model is a basic hand with only a single touch apparatus, but other developers may want to incorporate other features such as fully touchable skin or proprioception neurons.

\section{Evolving Simulations}

 Often times reinforcement learning algorithms are split into an environment and agent.  The environment contains the rules of the world such as the physics and and sensory input.  The agent is a program that resides in the environment and takes sensory data as input to be processed and then chooses an action. So the agent is an implicit part of the environment, but the the algorithms written for it reside on the agent side acting as the brain. The agent is typically where the most time is spent in algorithm development while the environment is usually treated as an abstract object to read from, but not to modify its source code. In the case of SenseNet, we have chosen to start with a simple environment that includes a hand. We know that our model of the hand is not a perfect representation of a human hand, as we only implemented a basic model of merkel cells at the tip of the index finger. Our strategy is to release new environments over time that incorporate more sophisticated and realistic models of sensorimotor systems and human skin with the goal of eventually having full humanoid bodies simulated. We follow the OpenAI convention of versioning every new environment we release to allow researchers to track their progress and test against the different environments. Moreover, the api has been designed to allow any researcher to create and submit new environments to share with the research community.

\section{SenseNet Benchmarks}

To measure progress, we have provided a baseline benchmark for researchers to collaborate and see if they can beat the performance of other researchers. Our benchmarks are meant for peer review and friendly competition. We want to encourage the sharing of ideas and code, not just who wrote the best algorithm.
With the initial release of SenseNet, we have included one benchmark we call "blind object classification".  It is similar to standard object classification where a model is trained on images from N number of classes. Then to measure performance, unseen images are given to the model and the model returns what is believes those image to be. The key difference being that we allow up to 500 time steps through the environment for the agent to touch and interact with the object to make a classification decision, making this a hybrid classification and reinforcement learning problem. We have provided simple example agents for the blind object classification benchmark written in pytorch and tensorflow, two popular deep learning research frameworks for python. While there have been some attempts at building a single algorithm for reinforcement learning to do classification\cite{rlclassification}, we opted to go with a simpler approach and train 2 concurrent networks. One reinforcement learning model based off the actor-critic algorithm is trained and rewarded when it touches an object. For the second model, we implemented a 2 layer convolutional neural network thats feeds into an LSTM that then feeds into a linear layer. We train it on the touch sensor's data to classify the object. As the agent steps through the environment, both models give feedback to the agent to get to the goal of classifying the object correctly.

We plan to incorporate many other benchmarks in the future for researchers to compare and build upon. Potential benchmarks are picking up objects or moving objects in the least amount of steps.  We also welcome ideas for new benchmarks from the community.

\section{Future Directions}

SenseNet is an ambitious open source project to help accelerate computational neuroscience and machine learning research in tactile and sensorimotor systems. We will continue to collect and create more data for researchers to use.  Our hope is that our dataset will grow to thousands of  classes totaling in over 10 million objects. We hope that our data can be used in robotics research, machine learning research, along with image processing as well.
As we grow the dataset, we will also continue development on the simulator environments.   We hope that other researchers can use SenseNet to develop their own ideas and algorithms for how the brain processes information and how the brain uses sensorimotor data.
Beyond the use cases we have outlined, we hope that researchers will come up with new and novel ways to use SenseNet.

\medskip

%Bibliographic references


\begin{thebibliography}{9}

\bibitem{demis}
Demis Hassabis, Dharshan Kumaran, Christopher Summerfield, Matthew Botvinick
\textit{Neuroscience-Inspired Artificial Intelligence}
2017

\bibitem{wordnet}
C. Fellbaum
\textit{WordNet: An Electronic Lexical Database.}
1998

\bibitem{imagenet}
Jia Deng et al.
\textit{ImageNet: A Large-Scale Hierarchical Image Database}
2009

\bibitem{alexnet}
Alex Krizhevsky, Ilya Sutskever, and Geoffrey E. Hinton
\textit{ImageNet Classification with Deep Convolutional Neural Networks}
2012

\bibitem{shapenet}
Angel Chang et al.
\textit{ShapeNet: An Information-Rich 3D Model Repository}
2015

\bibitem{ycb}
Calli et al
\textit{The YCB object and Model set: Towards common benchmarks for manipulation research.}
2015

\bibitem{rlclassification}
Marco Wiering et al.
\textit{Reinforcement Learning Algorithms for solving Classification Problems}
2011

\bibitem{pybullet}
Erwin Coumans and Yunfei Bai
\textit{pybullet, a Python module for physics simulation for games, robotics and machine learning}
2016

\bibitem{openai}
Greg Brockman et al.
\textit{OpenAI Gym}
2016

\bibitem{numpy}
Stéfan van der Walt, S. Chris Colbert and Gaël Varoquaux
\textit{The NumPy Array: A Structure for Efficient Numerical Computation, Computing in Science \& Engineering}
2011

\bibitem{skinpic}
Blausen.com staff
\textit{WikiJournal of Medicine}
2014

\bibitem{cldnn}
Tara N. Sainath, Oriol Vinyals, Andrew Senior, Hasim Sak
\textit{Convolutional, Long Short-Term Memory, Fully Connected Deep Neural Networks}
2015

\bibitem{attention}
Kelvin Xu, Jimmy Lei Ba,Ryan Kiros, Kyunghyun Cho, Aaron Courville, Ruslan Salakhutdinov, Richard Zemel, and Yoshua Bengio
\textit{Show, Attend and Tell: Neural Image Caption Generation with Visual Attention}
2016



\bibitem{proprio}
Matej Hoffman and Nada Bednarova
\textit{The encoding of proprioceptive inputs in the brain: knowns and unknowns from a robitic perspective}
2012



\bibitem{numenta}
Subutai Ahmad, Jeff Hawkins
\textit{Untangling Sequences: Behavior vs. External Causes}
2017






\end{thebibliography}
\end{document}